\pdfoutput=1

\documentclass[11pt]{article}

\usepackage{acl}
\usepackage{algorithm}
\usepackage{algpseudocode}

\usepackage{times}
\usepackage{latexsym}

\usepackage[T1]{fontenc}

\usepackage[utf8]{inputenc}

\usepackage{microtype}

\usepackage{array}
\usepackage{multirow}
\usepackage{amssymb}
\usepackage{amsmath}
\usepackage[pdftex]{graphicx}
\usepackage{subcaption}

\newcolumntype{P}[1]{>{\centering\arraybackslash}p{#1}}

\title{First the Worst: Finding Better Gender Translations During Beam Search}
\author{Danielle Saunders\thanks{ \hspace*{0.5em}Now at RWS Language Weaver} \and Rosie Sallis \and Bill Byrne \\
    Department of Engineering, University of Cambridge, UK  \\
      {\tt \{ds636, rs965\}@cantab.ac.uk, wjb31@cam.ac.uk}}

\begin{document}
\maketitle
\begin{abstract}
Generating machine translations via beam search seeks the most likely output under a model. However, beam search has been shown to amplify demographic biases exhibited by a model. We aim to address this, focusing on gender bias resulting from systematic errors in grammatical gender translation. Almost all prior work on this problem adjusts the training data or the model itself. By contrast, our approach changes only the inference procedure. 

We constrain beam search to improve gender diversity in n-best lists, and rerank n-best lists using gender features obtained from the source sentence. Combining these strongly improves WinoMT gender translation accuracy for three language pairs without additional bilingual data or retraining.   We also demonstrate our approach's utility for consistently  gendering  named entities, and its flexibility to handle new gendered language beyond the binary.
\end{abstract}

\section{Introduction}

Neural language generation models optimized by likelihood have a tendency towards  `safe' word choice. This lack of output diversity has been noted in  NMT \citep{vanmassenhove-etal-2019-lost} and throughout NLP \citep{li-etal-2016-diversity,sultan-etal-2020-importance}. Model-generated language may be repetitive or stilted. More insidiously, generating the most likely output based only on corpus statistics can amplify any existing biases in the corpus \citep{zhao-etal-2017-men}. 

Potential harms arise when biases around  word choice or grammatical gender inflections reflect demographic or social biases  \citep{sun-etal-2019-mitigating}.  The resulting gender mistranslations could involve implicit misgendering of a user or other referent, or  perpetuation of social stereotypes about the `typical' gender of a referent in a given context.

Past approaches to the problem almost exclusively involve retraining \citep{vanmassenhove-etal-2018-getting, escude-font-costa-jussa-2019-equalizing,bergmanis2020mitigating} or tuning  \citep{saunders-byrne-2020-reducing,basta-etal-2020-towards} on gender-adjusted data. Such approaches are often computationally expensive and risk introducing new biases \citep{shah-etal-2020-predictive}. Instead, we seek to improve translations from existing models.  \citet{roberts2020decoding} highlight beam search's tendency to amplify gender bias and \citet{renduchintala-etal-2021-gender} show that very shallow beams degrade gender translation accuracy; we instead guide beam search towards better gender translations further down the n-best list. 

Our contributions are as follows: we rerank NMT n-best lists, demonstrating that we can extract better gender translations from the \emph{original model's} beam. We also generate new n-best lists subject to gendered inflection constraints, and show this makes correctly gendered entities more common in n-best lists. We make no changes to the NMT model or training data, and require only monolingual resources for the source and target languages.

\subsection{Related work}
Prior work mitigating gender bias in NLP often involves adjusting training data,  directly \cite{zhao-etal-2018-gender} or via embeddings \cite{bolukbasi2016man}. Our inference-only approach is closer to  work on controlling or `correcting' gendered output.

Controlling gender translation generally involves introducing external information into the model. \citet{miculicich-werlen-popescu-belis-2017-using}  integrate cross-sentence coreference links into reranking to improve pronoun translation. \citet{vanmassenhove-etal-2018-getting} and \citet{moryossef-etal-2019-filling}  incorporate sentence-level gender features into training data and during inference respectively.  Token-level source gender tags are used by \citet{bergmanis2020mitigating} and \citet{saunders-etal-2020-neural}.    As in this prior work, our focus is applying linguistic gender-consistency information, rather than obtaining it.

A separate line of work treats gender-related inconsistencies as a search and correction problem. \citet{roberts2020decoding} find that beam search amplifies gender bias compared to sampling search. \citet{saunders-byrne-2020-reducing} rescore translations with a model fine-tuned for additional gender sensitivity,  constraining outputs to gendered-reinflections of the original. Related approaches for monolingual tasks  reinflect whole-sentence gender \citep{habash-etal-2019-automatic, alhafni-etal-2020-gender, sun2021they}. An important difference in our work is use of the same model for initial translation and reinflection,  reducing computation and complexity.

\section{Finding consistent gender in the beam}
There are two elements to our proposed approach. First, we \emph{produce an n-best list} of translations using our single model per language pair. We use either standard beam search or a two-pass approach where the second pass searches for differently-gendered versions of the highest likelihood initial translation. 
We then \emph{select a translation} from the list, either by log likelihood or by how far the target language gender features correspond to the source sentence.

\subsection{Gender-constrained n-best lists}
\label{ss:n-best}

\begin{figure}[ht]
    \centering
    \includegraphics[width=\linewidth]{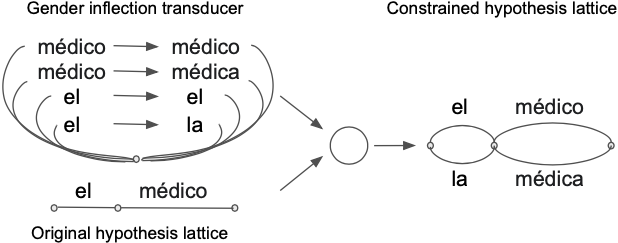}
    \caption{Constraints for a toy initial hypothesis.}
    \label{fig:lattice}
\end{figure}
We produce n-best lists in two ways. One option is standard  beam search. Alternatively, we synthesize n-best lists using the gendered constraint scheme of \citet{saunders-byrne-2020-reducing}, illustrated in  Figure \ref{fig:lattice}. This  involves a second \emph{gender-constrained} beam search pass to reinflect an initial hypothesis, producing a synthesized n-best list containing gendered alternatives of that hypothesis.

The second reinflection pass uses a target language \emph{gender inflection transducer} which defines grammatically gendered reinflections. 
For example, Spanish definite article \emph{el} could be unchanged or reinflected to \emph{la}, and profession noun \emph{médico} could  be reinflected to \emph{médica} (and vice versa). Composing the reinflections with the original hypothesis generates a \emph{constrained hypothesis lattice}.

We can now perform constrained beam search, which can encourage NMT to output specific vocabulary \citep{stahlberg-etal-2016-syntactically, khayrallah-etal-2017-neural}.
The only difference from standard beam search is that  gender-constrained search only expands translations forming paths in the constrained hypothesis lattice. In the Figure \ref{fig:lattice} example, beam-$n$ search would produce the $n$ most likely translations, while the gender-constrained  pass would only produce  the 4 translations  in the lattice. 

Importantly, for each language pair we use just one NMT model to produce gendered variations of its \emph{own}  hypotheses. Unlike \citet{saunders-byrne-2020-reducing} we do not reinflect translations with a separate gender-sensitive model. This removes the  complexity, potential bias amplification and computational load of developing the gender-translation-specific models central to their approach.

While we perform two full inference passes  to simplify implementation, further efficiency improvements are possible. For example, the source sentence encoding could be reused for the reinflection pass. In principle, some beam search constraints could be applied in the first inference pass, negating the need for two passes. These potential efficiency gains would not be possible if using a separate NMT model to reinflect the translations. 
\begin{figure*}[ht]
    \centering
    \includegraphics[width=\linewidth]{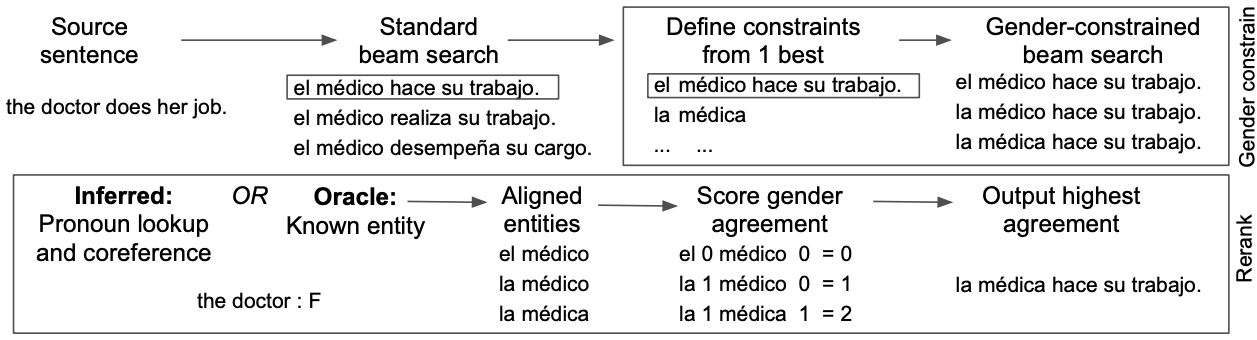}
    \caption{Complete workflow for a toy en-es example. We have two options for producing an n-best list - standard or gender-constrained search - and can then either take the highest likelihood output from the list, or rerank it.}
    \label{fig:workflow}
\end{figure*}
\subsection{Reranking gendered translations}
\label{ss:rerank}
\begin{algorithm}
\caption{Gender-reranking an n-best list}\label{alg:cap}
\textbf{Input:} $x$: Source sentence; $Y$: set of translation hypotheses for $x$; $L$: Log likelihoods for all $y \in Y$; $A$: word alignments between $x$ and all $y$ 
\begin{algorithmic}
\State $p, p_g \gets \text{pronoun\_and\_gender}(x)$ \Comment{Or oracle}
\State $e \gets \text{get\_entity}(x, p)$ \Comment{Or oracle}
\ForAll{$y \in Y$} 
\State $y_{score} \gets 0$
\ForAll{$t \in A_y(e)$}  \Comment{Translated entity}
 \State $t_g \gets \text{get\_gender(t)}$
 \If{$t_g = p_g$}
 \State $y_{score}\mathrel{+}= 1$
 \EndIf
\EndFor
\EndFor
\State $\hat{Y}=\{\text{argmax}_{y}(y_{score}, y \in Y)\}$
\State $\hat{y}=\text{argmax}_{y}(L(y), y \in \hat{Y})$\\
\Return $\hat{y}$
\end{algorithmic}
\end{algorithm}

We  select an output translation from an n-best list in two ways, regardless of whether the list was produced by beam search or the two-pass approach.  One option  selects the highest-likelihood translation under the NMT model. Alternatively, we rerank for  gender consistency with the source sentence. We focus on  either \emph{oracle} or \emph{inferred} entities  coreferent with a source pronoun. 

The \emph{oracle} case occurs in several scenarios. Oracle entity labels could be provided  as for the WinoMT challenge set \citep{stanovsky-etal-2019-evaluating}. They could also be user-defined for known entities \citep{vanmassenhove-etal-2018-getting}, or if translating the same sentence  with different entity genders to produce multiple outputs \citep{moryossef-etal-2019-filling}. 

The \emph{inferred} case determines  entities automatically given a source pronoun\footnote{In \ref{sec:names} we show this could also be a source named entity.} and its grammatical gender.  We  find coreferent entities  using a target language coreference resolution tool in get\_entity.  For brevity  Algorithm 1 is written for one entity per sentence: in practice there is no such limit.

For each entity we find the aligned translated entity, similar to \citet{bergmanis2020mitigating}. We determine the translated entity's grammatical gender by target language morphological analysis in get\_gender. Finally we rerank, first by source gender agreement, tie-breaking with log likelihood\footnote{Reranking code and n-best lists at \url{https://github.com/DCSaunders/nmt-gender-rerank}}.

\section{Experimental setup}
\label{ss:data}
We translate English into German, Spanish and Hebrew using Transformers \citep{vaswani2017attention}.
 We train the en-de model on WMT19 newstask data including filtered Paracrawl \cite{barrault-etal-2019-findings}, en-es on UNCorpus data \citep{ziemski-etal-2016-united}, and en-he on the IWSLT corpus \citep{cettolo2014report}.  For further training details see Appendix \ref{appendix-experimental}.

Some proposed steps require tools or  resources: 1) For gender-constrained search, creating gender inflection transducers;  2) For inferred-reranking, finding source gendered entities 3) For all reranking, finding translated gendered entities; 4) For all reranking, getting translated entity genders.

For 1) we use Spacy  \citep{honnibal2017spacy} and DEMorphy \citep{altinok2018demorphy} morphological analysis for Spanish and German, and fixed rules for Hebrew, on large vocabulary lists to produce gender transducers, following  \citet{saunders-byrne-2020-reducing}\footnote{Scripts and data for lattice construction as in \citet{saunders-byrne-2020-reducing} provided at \url{https://github.com/DCSaunders/gender-debias}}. The highest likelihood outputs from beam-4 search form the original hypothesis lattices. For 2) we use a RoBERTa model \citep{liu2019roberta} tuned for coreference on Winograd challenge data\footnote{Model from \url{https://github.com/pytorch/fairseq/tree/master/examples/roberta/wsc}}. For 3) we use fast\_align \citep{dyer-etal-2013-simple}. For 4) we use the same morphological analysis as in 1, now on translated entities.

We evaluate gender translation on WinoMT \citep{stanovsky-etal-2019-evaluating} via accuracy and $\Delta$G (F1 score difference between masculine and feminine labelled sentences, closer to 0 is better). As WinoMT lacks references we assess cased  BLEU on WMT18 (en-de), WMT13 (en-es) and IWSLT14 (en-he) using SacreBLEU 
 \citep{post-2018-call}.

\section{Results and discussion}

\begin{table*}[t]
    \centering
    \small 

    \begin{tabular}{p{0.05cm}|c|cc|ccc|ccc|ccc|}
\cline{2-13}
      &   \multirow{2}{*}{\textbf{Beam}} & \textbf{Gender}  & \textbf{Oracle}& \multicolumn{3}{c|}{\textbf{en-de}} &  \multicolumn{3}{c|}{\textbf{en-es}} & \multicolumn{3}{c|}{\textbf{en-he}}\\
       & & \textbf{constrain} & \textbf{rerank} &BLEU & Acc &  $\Delta$G & BLEU & Acc &   $\Delta$G& BLEU & Acc &   $\Delta$G\\
\cline{2-13}

  \footnotesize{1} &      \multirow{4}{*}{4}  & $\times$ & $\times$ &  \textbf{42.7} & 60.1 & 18.6 & 27.5 &  47.8& 38.4  & 23.8&47.5 &21.1\\
  \footnotesize{2} &         & \checkmark & $\times$  &  \textbf{42.7} & 59.1  & 20.1  & \textbf{27.8} & 48.3 & 36.2 & 23.8 & 47.4& 21.5 \\

   \footnotesize{3} &          & $\times$ & \checkmark & - & 66.5  &10.1   & - & 53.9 & 25.9 & - &52.0&16.8\\
   \footnotesize{4} &         & \checkmark& \checkmark & - & 77.9&\textbf{-0.6}& - &55.7& 22.3& -& 54.5& 13.7\\
\cline{2-13}

  \footnotesize{5} &             \multirow{4}{*}{20}  & $\times$ & $\times$ & 42.3 & 59.0 &  20.1 & 27.3 & 46.4 & 40.7 & \textbf{24.0} & 46.8&22.5\\
   \footnotesize{6} &          & \checkmark & $\times$ & \textbf{42.7} & 59.0 & 20.3& \textbf{27.8} & 48.3 & 36.2&  23.8 & 47.3 & 21.7\\
   \footnotesize{7} &          & $\times$ & \checkmark & - & 74.3 &2.4  & - &63.5&11.0   & - &59.3&11.2\\
  \footnotesize{8} &           & \checkmark & \checkmark & - & \textbf{84.2} & -3.6 & - &\textbf{66.3} & \textbf{8.1}  & - &\textbf{65.3}&\textbf{4.9}\\

\cline{2-13}
    \end{tabular}
    \caption{Accuracy (\%) and masculine/feminine F1 difference $\Delta$G, oracle-reranking WinoMT.  BLEU scores are for en-de WMT18, en-es WMT13, and en-he IWSLT14, which lack  gender labels so cannot  be oracle-reranked. 
    }
    \label{tab:mfresults}
\end{table*}

\begin{table*}[t]
    \centering
    \small 

    \begin{tabular}{p{0.05cm}|c|cc|ccc|ccc|ccc|}
                  \cline{2-13}
           &   \multirow{2}{*}{\textbf{Beam}} & \textbf{Gender}  & \textbf{Inferred}  & \multicolumn{3}{c|}{\textbf{en-de}} &  \multicolumn{3}{c|}{\textbf{en-es}} & \multicolumn{3}{c|}{\textbf{en-he}}\\
   & & \textbf{constrain} & \textbf{rerank}  &BLEU & Acc &  $\Delta$G & BLEU& Acc &   $\Delta$G& BLEU& Acc &   $\Delta$G\\
                  \cline{2-13}

    \footnotesize{1}&      \multirow{2}{*}{4}  & $\times$  & \checkmark  &42.7 & 65.9   & 10.7  & 27.5 &52.6  &28.1  &23.8&51.3&17.0 \\
      \footnotesize{2}  &   & \checkmark& \checkmark &42.7 & 76.4  & 0.5  & 27.8 & 53.9  & 24.6  & 23.8 &53.6&14.4  \\

                  \cline{2-13}
       \footnotesize{3}&     \multirow{2}{*}{20} & $\times$  & \checkmark& 42.2& 72.9  &3.3   & 27.3& 60.2 & 15.3 &24.0 &57.8& 11.9\\
     \footnotesize{4} &     & \checkmark & \checkmark &42.6  & 81.8 & -2.6 & 27.8& 63.5& 10.9&23.8 &62.8& 6.2\\
                  \cline{2-13}
    \end{tabular}
    \caption{Accuracy (\%) and masculine/feminine F1 difference $\Delta$G. Inferred-reranking with genders and entities for WinoMT and generic test sets determined by a  RoBERTa model. Non-reranked results unchanged from Table \ref{tab:mfresults}.}
    \label{tab:roberta}
\end{table*}

\subsection{Oracle entities}
We first describe oracle-reranking n-best lists in Table \ref{tab:mfresults}, before proceeding to the more general scenario of inferred-reranking.  Comparing lines 1 vs 2, gender-constrained beam-4 search taking the highest likelihood output scores similarly to standard beam-4 search for all metrics and language pairs. For beam-20 (5 vs 6) en-de and en-es, constraints do mitigate the  BLEU degradation common with larger beams  \citep{stahlberg-byrne-2019-nmt}. 




In lines 1 vs 3, 5 vs 7, we oracle-rerank beam search outputs instead of choosing by highest likelihood. We see about 10\%  accuracy improvement relative to non-reranked beam-4 across languages, and over 25\% relative improvement  for beam-20. Combining oracle-reranking and constraints further boosts accuracy. This suggests constraints encourage presence of better gender translations in n-best lists, but that reranking is needed to extract them.

Using beam-20 significantly improves the performance of  reranking. With constraints, beam-20 oracle-reranking gives \emph{absolute} accuracy gains of about 20\% over the highest likelihood beam search output. However, beam-4 shows most of the improvement over that baseline. We find diminishing returns as beam size increases (Appendix \ref{appendix-beamsizes}), suggesting  large, expensive beams are  not necessary.

\subsection{Inferred entities}

We have shown accuracy improvements with oracle reranking, indicating that the synthesized n-best lists often contain a gender-accurate hypothesis. In Table \ref{tab:roberta}, we explore inferred-reranking using a RoBERTa model, investigating whether that hypothesis can be found automatically. We find very little degradation in WinoMT accuracy when inferring entities  compared to the oracle  (Table \ref{tab:mfresults}). We hypothesise that difficult sentences  are hard for both coreference resolution and NMT, so cases where RoBERTa disambiguates wrongly are also  mistranslated with oracle information.

We are unable to oracle-rerank the generic test sets, since they have no oracle gender labels. However, we can tag them using RoBERTA for inferred-reranking. In Table \ref{tab:roberta} we find this has little or no impact on BLEU score, unsurprising for sets not designed to highlight potentially subtle gender translation effects. This suggests positively that our scheme does not impact general translation quality.

\begin{table}[t]
    \centering
    \small 

    \begin{tabular}{|c|c|ccc|}
    \hline
         \textbf{Beam} & \textbf{System} & \textbf{en-de} &  \textbf{en-es} & \textbf{en-he}\\
                  \hline
 \multirow{2}{*}{4}  &S\&B &   79.4 & 62.2  & 53.1  \\
& S\&B + rerank & 81.9 & 68.9 &  56.6 \\

    \hline
 \multirow{2}{*}{20}  & S\&B & 79.6  &62.1& 52.8  \\
& S\&B + rerank & 83.6 & 73.9 & 62.9 \\\hline
    \end{tabular}
    \caption{WinoMT accuracy inferred-reranking the adaptation scheme of \citet{saunders-byrne-2020-reducing}.}
    \label{tab:sandb}
\end{table}
\begin{table*}[ht]
    \centering
    \begin{small}
    \begin{tabular}{|c|p{12cm}|}
    \hline
   \multicolumn{2}{|p{13cm}|}{Vallejo appears to have only narrowly edged out Calderon, \textbf{who} had led polls before election day} \\
\hline
-12.3 &Vallejo scheint nur knapp ausgegrenzt Calderon, \textbf{der} vor dem Wahltag Wahlen geführt hatte. \\
-14.6 & $\ast$ Vallejo scheint nur knapp ausgegrenzt Calderon, \textbf{die} vor dem Wahltag Wahlen geführt hatte.\\
-24.3 & Vallejo scheint nur knapp ausgegrenzt Calderon, \textbf{der} vor dem Wahltag Wahlern geführt hatte.\\
-26.5 & Vallejo scheint nur knapp ausgegrenzt Calderon, \textbf{die} vor dem Wahltag Wahlern geführt hatte.\\
\hline
    \end{tabular}
    \caption{Sentence from WMT newstest12 with gender-constrained n-best list and NLL scores. Words like `who' coreferent with `Calderon'  become entities for Algorithm 1, which finds a better gendered translation ($\ast$).}
    \label{tab:nonpronounexamples}
    \end{small}
\end{table*}
So far we have not changed the NMT model at all. In Table \ref{tab:sandb}, for comparison, we investigate the approach of \citet{saunders-byrne-2020-reducing}: tuning a model on a  dataset of gendered profession sentences, then constrained-rescoring the original model's hypotheses.\footnote{Different scores from the original work may be due to variations in hyperparameters, or WinoMT updates.} We do indeed see strong gender accuracy improvements with this approach, but inferred-reranking the resulting models' n-best lists further improves scores.  We also note that inferred reranking the baseline with beam size 20 (Table \ref{tab:roberta} line 4) outperforms non-reranked  S\&B, without requiring specialized profession-domain tuning data or any change to the model.

\subsection{Reranking with named entities}
\label{sec:names}

At time of writing, published gender translation test sets focus on profession nouns, a domain we evaluate with WinoMT. 
However, our approach can also improve other aspects of gender translation. One of these is consistently gendering named entities. Sentences  may contain gendered terminology with no pronouns, only named entities. Generic name-gender mappings are unreliable: many names are not  gendered, and a  name with a `typical' gender may not correspond to an individual's  gender. However, we may know the appropriate gendered terms to use for a \emph{specific} named entity, for example from other sentences, a knowledge base, or user preference. With this information we can gender-rerank.

An example is given in Table \ref{tab:nonpronounexamples}. The English sentence   contains no gendered pronoun, so is not covered by our default reranking algorithm. We know from previous sentences that Calderon should be referred to with the linguistic feminine, so we can rerank with known $p_g$. The `entities' $e$ are the words referring to Calderon, including `who', `had' and `led'.\footnote{Extracted using  RoBERTa coreference model; future work might explore use of a lightweight dependency parser.} Algorithm 1 proceeds over these entities, of which only `who' is gendered in German, to extract a better gendered translation.

\subsection{Reranking with new gendered language}

Another benefit of our approach is flexibility to introducing new gendered vocabulary, e.g. as used by non-binary people.  Developing a system to correctly produce new terms like neopronouns  is itself an open research problem \citep{saunders-etal-2020-neural}. However, we can simulate such a system by editing existing WinoMT translations to contain  gendered-term placeholders instead of binary gendered terms, and shuffling these translations into n-best lists. For example, where a German translation includes \emph{der Mitarbeiter}, the  employee (masculine), we substitute \emph{DEFNOM MitarbeiterNEND}. This allows later replacement of \emph{DEFNOM} by e.g. \emph{dier} or \emph{NEND} by \emph{\_in} \citep{heger2020xier}, but remains flexible to preferences for new gendered language.
We then  define the new patterns for identification by the reranker.

To evaluate reranking with new gendered language, we use 1826 neutral WinoMT sentences with they/them pronouns on the English side. We initialise the corresponding n-best lists with the masculine WinoMT German 20-best lists, and shuffle one  `placeholder'  translation into each, giving them the average log likelihood of the whole list. We find that the reranker successfully extracts the correct placeholder-style sentences in 92\% of cases. This demonstrates that if a system can generate some new gendered term, reranking can extract it from an n-best list with minimal adjustments.


\section{Conclusions}
This paper attempts to improve gender translation without a single change to the NMT model. We demonstrate that gender-constraining the target language during inference can encourage models to produce n-best lists with correct hypotheses. Moreover, we show that simple reranking heuristics can  extract more accurate gender translations from the n-best lists using oracle or inferred information.

Unlike other approaches to this problem we do not attempt to counter unidentified and potentially intractable sources of bias in the training data, or produce new models. However, our approach does significantly boost the accuracy of a prior data-centric bias mitigation technique. In general we view our scheme as orthogonal to such approaches: if a model ranks diverse gender translations higher in the beam initially, finding better gender translations during beam search becomes simpler.

  \section*{Acknowledgments}
 This work was supported by EPSRC grants EP/M508007/1 and EP/N509620/1 and performed using resources from the Cambridge Tier-2 system operated by the University of Cambridge Research Computing Service\footnote{\url{http://www.hpc.cam.ac.uk}} funded by EPSRC Tier-2 capital grant EP/P020259/1. 

\section*{Impact statement}

Where machine translation is used in people's lives, mistranslations have the potential to misrepresent people. This is the case when personal characteristics like social gender conflict with model biases towards certain forms of grammatical gender. As mentioned in the introduction, the result can involve implicit misgendering of a user or other human referent, or  perpetuation of social biases about gender roles as represented in the translation. A user whose words are translated with gender defaults that imply they hold such biased views will also be misrepresented.

We attempt to avoid these failure modes  by identifying translations which are at least consistent within the translation and consistent with the source sentence. This is dependent on identifying grammatically gendered terms in the target language -- however, this element is very flexible and can be updated for new gendered terminology. We note that models which do not account for variety in gender expression such as neopronoun use may not be capable of generating appropriate gender translations. However, we demonstrate that, if definable, a variety of gender translations can be extracted from the beam.

By avoiding the data augmentation, tuning and retraining elements in previously proposed approaches to gender translation, we  simplify the process and remove additional stages where bias could be introduced or amplified \citep{shah-etal-2020-predictive}.

In terms of compute time and power, we minimize impact by using a single GPU only for training the initial NMT models exactly once for the iterations listed in Appendix \ref{appendix-experimental}. All other experiments involve inference or rescoring the outputs of those models and run in parallel on CPUs in under an hour, except the experiments following \citet{saunders-byrne-2020-reducing}, an approach itself involving only minutes of GPU fine-tuning.

\bibliographystyle{acl_natbib}
\bibliography{refs}

\clearpage

\appendix
\section{Model training details}
\label{appendix-experimental}
All NMT models are 6-layer Transformers with 30K BPE vocabularies \citep{sennrich-etal-2016-neural}, trained using Tensor2Tensor with batch size 4K \citep{vaswani-etal-2018-tensor2tensor}.
All data except Hebrew is truecased and tokenized using \citep{koehn-etal-2007-moses}. The en-de model is trained for 300K batches, en-es  for 150K batches, and en-he for 15K batches, transfer learning from the en-de model.  We filter subworded data for max (80) and min (3) length, and  length ratio 3. We evaluate cased  BLEU on WMT18 (en-de, 3K sentences), WMT13 (en-es, 3K sentences) and IWSLT14 (en-he, 962 sentences). For validation during NMT model training we use earlier test sets from the same tasks.

\section{Beam size for constrained reranking}
\label{appendix-beamsizes}
In this paper we present results with beam sizes 4 and 20. Beam-4 search is  commonly used and meets a speed-quality trade-off for NMT (see e.g. \citet{speedquality2016}). Beam-20 is still practical, but approaches  diminishing returns for quality without search error mitigation \citep{stahlberg-byrne-2019-nmt}. These sizes therefore  illustrate contrasting levels of practical reranking.  However, it is instructive to explore what beam size is necessary to benefit from gender-constrained reranking. 

\begin{figure}[ht]
    \centering
    \includegraphics[width=0.84\linewidth]{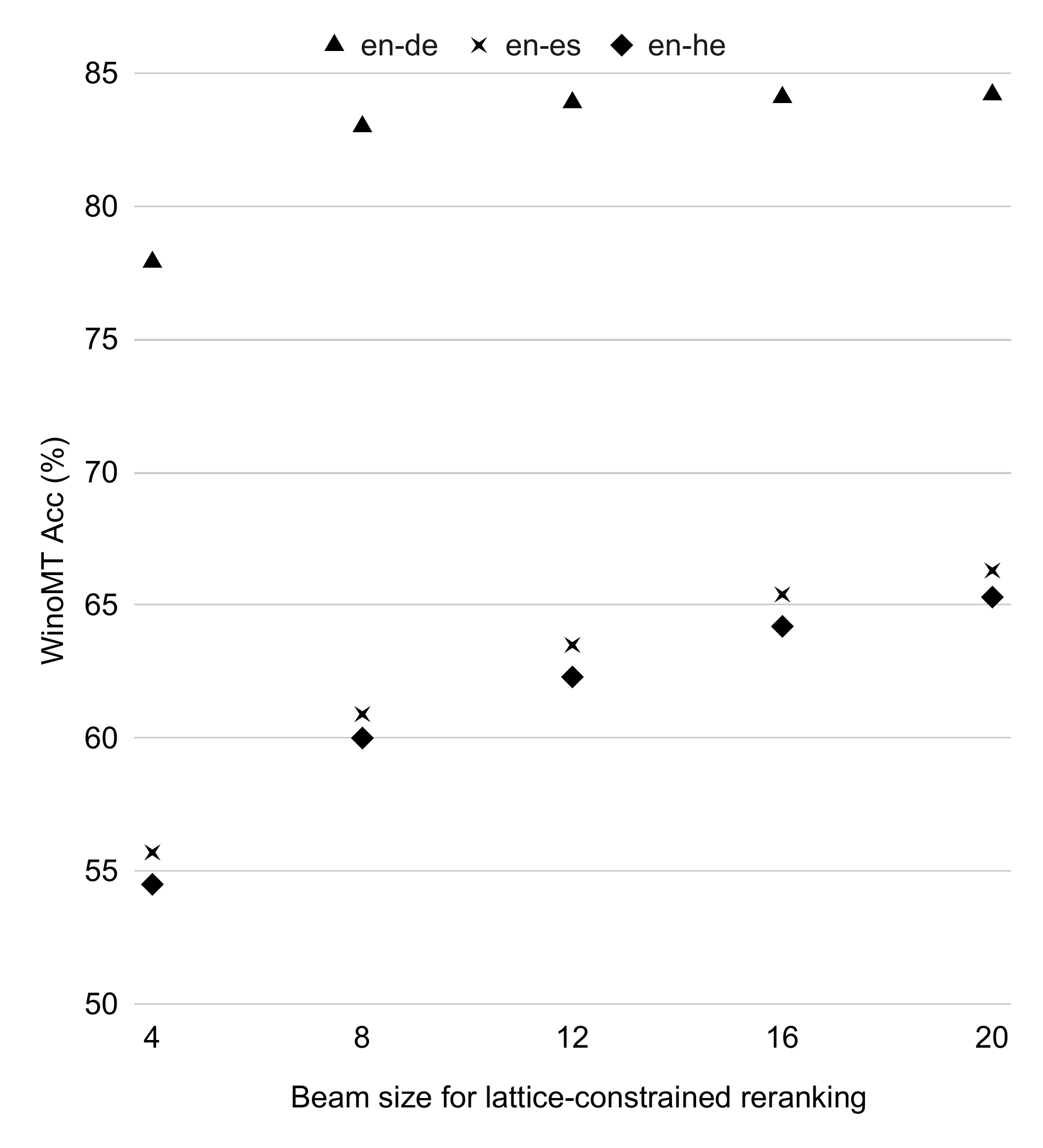}
    \caption{WinoMT accuracy after oracle-reranking gender-constrained n-best lists, varying n.}
    \label{fig:beamsize}
\end{figure}

In Figure \ref{fig:beamsize} we report WinoMT accuracy under gender-constrained oracle reranking with beam width increasing by intervals of 4. For all systems, the largest jump in improvement is between beam sizes 4 and 8, with diminishing returns after beam-12. The en-de curve is relatively shallow, possibly due to strong scores before reranking, or even a performance ceiling determined by the WinoMT framework itself. Curves for en-he and en-es are very close, suggesting a similarity between the gender distribution in the n-best lists for those models.

\section{Constrained vs unconstrained beams}
\label{appendix-beamexamples}
\begin{table*}[ht]
    \centering
    \begin{small}
    \begin{tabular}{|p{2.2 cm}|p{13cm}|}
    \hline
   \multicolumn{2}{|l|}{1. \underline{The broker} called the client because \underline{she} had new stocks to recommend.} \\
    \hline    \multirow{4}{*}{Unconstrained}     &  *	\emph{Der Makler} rief \emph{den Kunden} an, weil \underline{sie} neue Aktien zu empfehlen hatte.\\
& \emph{Der Makler} rief \underline{die Klientin} an, weil \underline{sie} neue Aktien zu empfehlen hatte. \\
& \emph{Der Makler} rief \emph{den Klienten} an, weil \underline{sie} neue Aktien zu empfehlen hatte.\\
&	\emph{Der Makler} rief \emph{den Kunden} an, weil \emph{er} neue Aktien zu empfehlen hatte.\\

    \hline
        \multirow{4}{*}{Constrained}     & \emph{Der Makler} rief \emph{den Kunden} an, weil \underline{sie} neue Aktien zu empfehlen hatte.  \\
&\emph{Der Makler} rief \emph{den Kunden} an, weil \emph{er} neue Aktien zu empfehlen hatte. \\
&*	\underline{Die Maklerin} rief \emph{den Kunden} an, weil \underline{sie} neue Aktien zu empfehlen hatte. \\
&\emph{Der Makler} rief \underline{die} \emph{Kunden} an, weil \underline{sie} neue Aktien zu empfehlen hatte.  \\

    \hline

        \multicolumn{2}{|l|}{2.  The teacher explained the situation to \underline{the chief} and felt understood by \underline{her}.} \\
    \hline
    \multirow{4}{*}{Unconstrained}     &   \underline{Die Lehrerin} erklärte die Situation \emph{dem Chef} und fühlte sich von \underline{ihr} verstanden. \\
&*	\underline{Die Lehrerin} erklärte \emph{dem Obersten} die Situation und fühlte sich von \underline{ihr} verstanden. \\
&	\underline{Die Lehrerin} erklärte die Situation \emph{dem Obersten} und fühlte sich von \underline{ihr} verstanden.\\
&	\underline{Die Lehrerin} erläuterte die Situation \emph{dem Obersten} und fühlte sich von \underline{ihr} verstanden.\\

      \hline

    \multirow{4}{*}{Constrained}     &*\underline{Die Lehrerin} erklärte die Situation \emph{dem Chef} und fühlte sich von \underline{ihr} verstanden. \\
&\emph{Der Lehrer} erklärte die Situation \emph{dem Chef} und fühlte sich von \underline{ihr} verstanden. \\
&\emph{Der Lehrer} erklärte die Situation \emph{dem Chef} und fühlte sich von \emph{ihm} verstanden.\\
&	\underline{Die Lehrerin} erklärte die Situation \emph{dem Chef} und fühlte sich von \emph{ihm} verstanden. \\
\hline
    \multicolumn{2}{|l|}{3. The assistant hired \emph{the driver} as \emph{he} offered an excellent service.} \\
    \hline
    \multirow{4}{*}{Unconstrained}     &   *	\emph{Der Assistent} stellte  \emph{den Fahrer} ein, da \emph{er} einen ausgezeichneten Service bot. \\
&\emph{Der Assistent} stellte  \emph{den Fahrer} ein, da \emph{er} einen exzellenten Service bot. \\
&	\emph{Der Assistent} stellte  \emph{den Fahrer} ein, da \emph{er} einen hervorragenden Service bot.\\
&\emph{Der Assistent} stellte  \emph{den Fahrer} ein, als \emph{er} einen ausgezeichneten Service bot.\\
      \hline

    \multirow{4}{*}{Constrained}     &*	\emph{Der Assistent} stellte \emph{den Fahrer} ein, da \emph{er} einen ausgezeichneten Service bot. \\
&	\underline{Die Assistentin} stellte  \emph{den Fahrer} ein, da \emph{er} einen ausgezeichneten Service bot. \\
&	\emph{Der Assistent} stellte \underline{die} \emph{Fahrer} ein, da \emph{er} einen ausgezeichneten Service bot.\\
&	\emph{Der Assistent} stellte  \emph{den Fahrer} ein, da \emph{er} eine ausgezeichnete Service bot. \\
    \hline

    \multicolumn{2}{|l|}{4.  \emph{The laborer} handed the application to the editor because \emph{he} wants the job.}\\
    \hline
    \multirow{4}{*}{Unconstrained}     &  *	\emph{Der Arbeiter} übergab den Antrag an \emph{den Herausgeber}, weil \emph{er} den Job will. \\
&		\emph{Der Arbeiter} übergab die Bewerbung an \emph{den Herausgeber}, weil \emph{er} die Arbeit will. \\
&		\emph{Der Arbeiter} übergab die Bewerbung an \emph{den Herausgeber}, weil \emph{er} den Job will.\\
&		\emph{Der Arbeiter} überreichte die Bewerbung an \emph{den Herausgeber}, weil \emph{er} die Arbeit will. \\

      \hline

    \multirow{4}{*}{Constrained}     &*	\emph{Der Arbeiter} übergab den Antrag an \emph{den Herausgeber}, weil \emph{er} den Job will.  \\
&	\emph{Der Arbeiter} übergab den Antrag an \emph{den Herausgeber}, weil \emph{er} diesen Job will.  \\
&	\emph{Der Arbeiter} übergab den Antrag an \emph{den Herausgeber}, weil \emph{er} die Job will.  \\
&\emph{Der Arbeiter} übergab die Antrag an \emph{den Herausgeber}, weil \emph{er} die Job will.  \\

    \hline
   \multicolumn{2}{|l|}{5. \emph{The developer} was unable to communicate with the writer because \emph{he} only understands the code.} \\
    \hline
    \multirow{4}{*}{Unconstrained}     &   *	\emph{Der Entwickler} konnte nicht mit \emph{dem Schriftsteller} kommunizieren, weil \emph{er} nur den Code versteht.\\
&	\emph{Der Entwickler} konnte nicht mit \emph{dem Autor} kommunizieren, weil \emph{er} nur den Code versteht. \\
&	\emph{Der Entwickler} war nicht in der Lage, mit \emph{dem Schriftsteller} zu kommunizieren, weil \emph{er} nur den Code versteht. \\
&	\emph{Der Entwickler} war nicht in der Lage, mit \emph{dem Autor} zu kommunizieren, weil \emph{er} nur den Code versteht.\\

      \hline

    \multirow{4}{*}{Constrained}     &*	\emph{Der Entwickler} konnte nicht mit \emph{dem Schriftsteller} kommunizieren, weil \emph{er} nur den Code versteht. \\
&	\emph{Der Entwickler} konnte nicht mit \underline{der Schriftstellerin} kommunizieren, weil \emph{er} nur den Code versteht. \\
&	\emph{Der Entwickler} konnte nicht mit \emph{dem Schriftsteller} kommunizieren, weil \emph{er} nur die Code versteht. \\
&	\emph{Der Entwickler} konnte nicht mit \emph{dem Schriftsteller} kommunizieren, weil \emph{er} nur diesen Code versteht.  \\

\hline
    \end{tabular}
    \caption{English-German 4-best lists for 5 randomly-selected WinoMT sentences, translated with normal beam search and gender-constrained beam search. Grammatically feminine human entities are \underline{underlined}.  Grammatically masculine human entities are \emph{emphasised}. Lists are ordered by NMT model likelihood (first is 1best) -  lines marked with * are those selected under oracle-reranking.\\
    1: Constrained reranking finds a better gender translation that is not present in the unconstrained beam.\\
    2: A better gendered translation is not found in either width-4 beam. Constraints still maintain semantic meaning throughout the beam while allowing syntactic variation, including a differently gendered secondary entity.\\
    3, 4, 5: The highest likelihood output is acceptable. For 3 and 5 constraining the n-best list results in more gender variation. }
    \label{tab:beamexamples}
    \end{small}
\end{table*}
We can observe the difference between standard and constrained beam search by examining the n-best lists. Table \ref{tab:beamexamples} (next page) gives 5 examples of 4-best lists for WinoMT sentences translated into German. Examples are not cherry-picked but selected from throughout WinoMT with a random number generator. Lists are ordered by NMT model likelihood and produced with standard unconstrained beam search, and with constrained beam search.

With standard beam search, translations vary  words unrelated to the entities, such as synonyms or verb tenses. However, entity grammatical genders are generally unchanged throughout the unconstrained n-best lists, except for 1 where the secondary entity changes. Reranking cannot always find a gender-consistent translation in the unconstrained lists, defaulting to the 1best for all except 2 (which seems to have a poorly aligned hypothesis).

By contrast, constrained beam search ensures the n-best list contains gendered inflections of the initial best-scoring translation. The changes  vary the grammatical genders of articles and entities, resulting in more gender-diverse hypotheses, and allowing reranking to find a better translation for 1.

We note that in 3, 4 and 5 both the pronoun and the default gender convention for unknown gender entities are masculine. Reranking is not strictly necessary to find a better translation for these sentences, since the highest likelihood output is gender-consistent. However, we note that some outputs with gender constraints  do in fact vary the gender of the secondary entity -- the entity  with unspecified gender. This illustrates our approach's ability to improve n-best list diversity even when it does not necessarily impact translation consistency. 

We observe occasional grammatical inconsistencies in n-best hypotheses (e.g. "die Fahrer" in 3). When constraining beam search to grammatical variations of a sentence with an imperfect NMT model, we expect some hypotheses with grammatical degradation. However, our priority, and the purpose of our reranking scheme, is  consistency with the source in the output translation, not inconsistencies elsewhere in the n-best list. 

\end{document}